\documentclass[10pt,twocolumn,letterpaper]{article}

\usepackage[pagenumbers]{cvpr} 

\usepackage{graphicx}
\usepackage{amsmath}
\usepackage{amssymb}
\usepackage{booktabs}
\usepackage{multirow}

\usepackage[pagebackref,breaklinks,colorlinks]{hyperref}

\usepackage[capitalize]{cleveref}
\crefname{section}{Sec.}{Secs.}
\Crefname{section}{Section}{Sections}
\Crefname{table}{Table}{Tables}
\crefname{table}{Tab.}{Tabs.}

\def\cvprPaperID{5156} 
\def\confName{CVPR}
\def\confYear{2022}

\begin{document}
\newcommand{\method}{SDG-SOD}
\newcommand{\dataset}{COCO SOD Annotation}
\renewcommand{\thefootnote}{\fnsymbol{footnote}}
\title{Semantic Distillation Guided Salient Object Detection}

\author{Bo Xu\textsuperscript{1}\thanks{Equal contribution.} , Guanze Liu\textsuperscript{1$*$}, Han Huang\textsuperscript{1}, Cheng Lu\textsuperscript{2},  Ziwen Li\textsuperscript{1} and Yandong Guo\textsuperscript{1}\thanks{The corresponding author.}\\
\textsuperscript{1}OPPO Research Institute, \textsuperscript{2}Xmotors\\
}
\maketitle

\begin{abstract}
   Most existing CNN-based salient object detection methods can identify local segmentation details like hair and animal fur, but often misinterpret the real saliency due to the lack of global contextual information caused by the subjectiveness of the salient object detection (SOD) task and the locality of convolution layers. Moreover, due to the unrealistically expensive labeling costs, the current existing SOD datasets are insufficient to cover the real data distribution. The limitation and the bias of the training data add additional difficulty to fully explore the semantic association between object-to-object and object-to-environment in a given image. In this paper, we propose a semantic distillation guided SOD (\method) method that produces accurate results by fusing semantically distilled knowledge from generated image captioning into the Vision-Transformer-based SOD framework. \method~can better uncover inter-objects and object-to-environment saliency and cover the gap between the subjective nature of SOD and its expensive labeling. Comprehensive experiments on five benchmark datasets demonstrate that the \method~outperforms the state-of-the-art approaches on four evaluation metrics, and largely improves the model performance on DUTS, ECSSD, DUT-OMRON, HKU-IS, and PASCAL-S datasets. The code will be released soon.
   \vspace{-9pt}
\end{abstract}

\section{Introduction}
\label{sec:intro}

Salient object detection is a basic computer vision task that aims to segment the objects in an image that attracts human attention~\cite{qin2019basnet}. 
Recent studies utilize Convolutional Neural Networks (CNN) for salient object detection and achieve remarkable results~\cite{ke2022recursive,qin2019basnet,qin2020u2,tang2021disentangled,wei2020f3net,wu2021decomposition,zhou2020interactive}. CNN-based models typically adopt an encoder-to-decoder architecture to fuse multi-scale features which is essential to grasp the salient contextual information. At the same time, such models typically achieve good granularity by establishing skip connections between encoder and decoder, which greatly improves the edge performance on a single object.

\begin{figure}[t]
    \centering
    \includegraphics[width=0.93\linewidth]{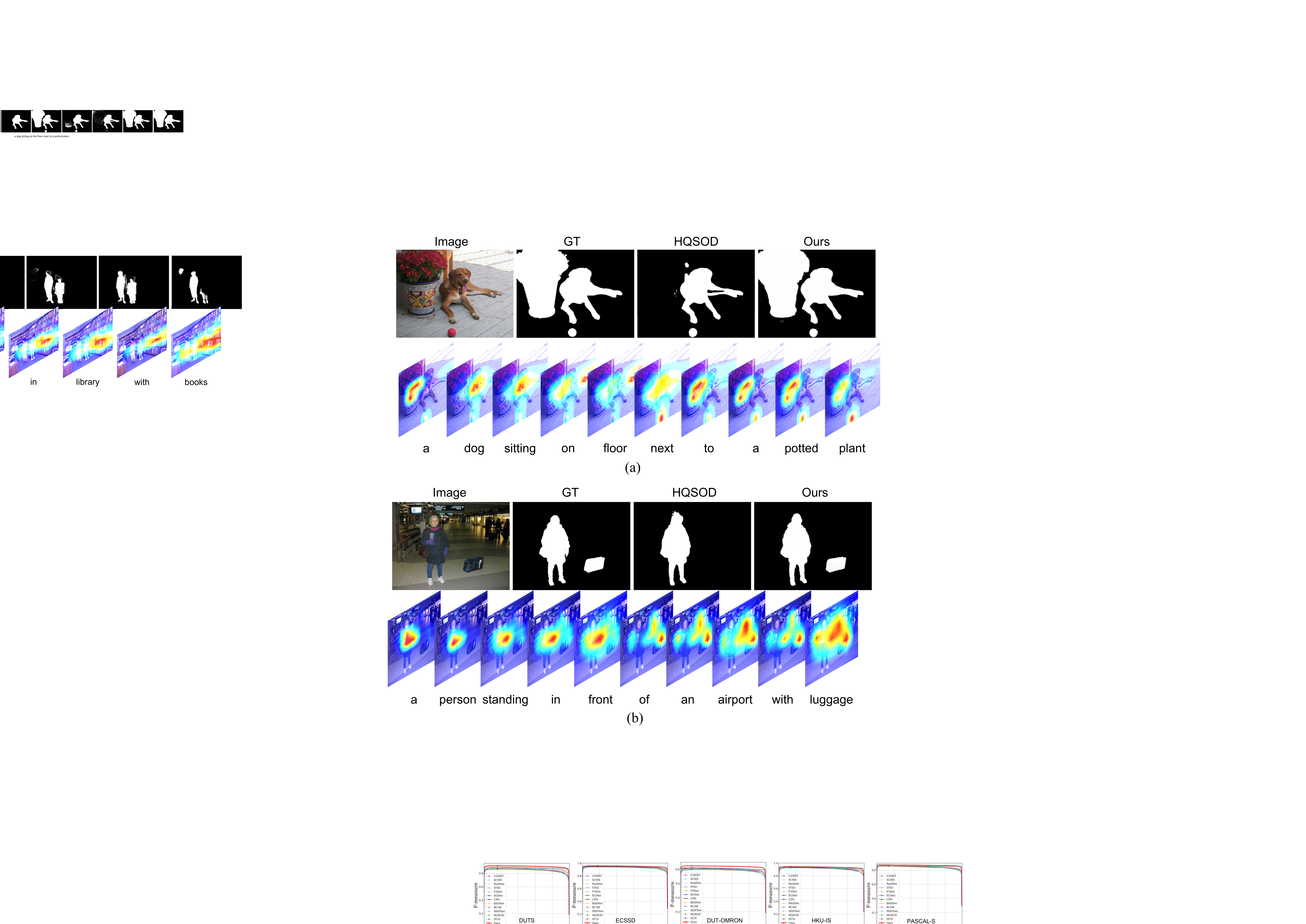}
    \caption{Given challenging SOD images, one of the recent state-of-the-art HQSOD~\cite{tang2021disentangled}, fails to predict fully salient objects. However, our \method~is able to capture accurate saliency information with the guidance of the generated image captioning.}
    \label{fig:introduction}
\vspace{-10pt}
\end{figure}

However, there still remain three big challenges in SOD prediction. First, unlike some well-defined tasks such as object detection or segmentation, the definition of saliency in an image is highly subjective. If we present the same image to a group of subjects, it's very natural that each individual will interpret this image a bit differently based on his or her educational background, gender, ethnic, or religious beliefs. Such variation impacts SOD labeling in many ways. For example, as shown in Figure~\ref{fig:introduction} (a), labeler A is more likely to interpret the flowers as the SOD objects together with the dog in this image, while labeler B only likes the pet. More examples about the subjective bias of SOD annotations are described in Appendix~\ref{limitation}. So the true distribution of such variation can only be learned by covering a large number of labeler populations. Unfortunately, since the SOD labeling requires pixel-level granularity, it is unrealistically expensive to acquire such a representative dataset to cover the real data distribution. Recent existing SOD training sets like DUTS~\cite{wang2017learning} are examples of such data insufficiency. Consequently, SOD models that trained from scratch often suffer from severe overfitting problems and fail to perform well on complex scenes. 

Second, even when sometimes the existing SOD models can find more than one salient object, they fail to extract completed and meaningful saliency due to insufficient understanding of the semantic association between a)object to object, b)object to the ambient environment, especially when salient objects are spatially far apart. One reason for that is previous methods often utilize an ImageNet pre-trained classification backbone, which lacks the ability to semantically correlate objects in a given image. More specifically, most of the backbone knowledge used in SOD prior arts are transferred from detection or segmentation tasks, which is designed only to single out the individual object. Its ability to correlate two or even more salient objects, especially far apart ones, and understand the global context is heavily limited by the receptive field of CNN. One example is in Figure~\ref{fig:introduction} (b). Although distant from the person, that unattended luggage, which is likely to cause suspicion for security reason, is undoubtedly salient given its airport location. 

Third, CNN models can generate fine-grained detailed local saliency prediction like animal fur, they require deeper layers to achieve larger receptive fields, which leads to an inevitable structure loss when the salient object is occluded or shows very strong shape or color variation. One example is shown in Row 3 of Figure~\ref{fig:visualization_pvt}, that a frog is partially occluded by a branch, the CNN-based methods (\ie (e) to (h)) cannot achieve good SOD performance due to the inability of CNN to reach a balance between local texture details and salient structural integrity.  

To address the above challenges, we propose the solution with the following three components accordingly. First, we introduce semantic distillation from the generated image caption as an extra modality branch that can guide the SOD model to capture more comprehensive saliency under a limited amount of supervision and labeling effort. That is inspired by the facts as follows: (a) The ground truth caption of one image commonly consists of descriptions from multiple individuals, which helps mitigate demographic bias, unlike SOD labeling. (b) In an image captioning sequence, subject and object, usually nouns, both explicitly hint the saliency in an image, so that it can cover all salient objects in an image more easily. (c) Compared with SOD pixel labeling, a large quantity of image captioning ground truth can be quickly collected at a low cost, since it takes only a few seconds for an individual to describe an image. Moreover, predicates (\eg verbs) and prepositions can correlate objects and establish environmental awareness, constructing associated structural information of this image in the image captioning network. Therefore, we believe transferring the above characteristics of image captioning to SOD can greatly improve its generalization power at minimum cost.  

Second, we introduce a Pyramid Vision Transformer~\cite{wang2021pyramid} (PVT) based network instead of a pure convolutional structure for saliency map prediction. We believe that the PVT-based backbones\cite{mao2021transformer} combine the advantages of both CNN and Transformer. Different from the local receptive of CNN, PVT produces a global receptive field that is suitable to maintain the structural completeness of the salient object. With a better global understanding, the model can better recognize the object that is occluded, irregular, or out of image boundary. Echoing the fact that the transformer-based method has proved to be the de facto SOTA on image captioning task\cite{desai2021virtex}, applying its pyramid variant on the SOD branch can organically fuse the visual features with their image captioning counterparts at multi-scale. In addition, PVT reduces the computations of large feature maps, different from Vision Transformer (ViT)~\cite{dosovitskiy2021an} which incurs high computational and memory costs.

We also propose a pre-training strategy to improve the performance of our PVT-based SOD model. We relabel the foreground classes on the COCO panoptic segmentation dataset and produce a novel COCO annotation called \dataset~with more than 30000 images. \dataset~is labeled in a way that is more suitable for saliency prediction. 

%
%

The \method~network consists of two branches: Semantically Distilled Guidance (SDG) and PVT-based Salient Object Detection (PVT-SOD). We first train the dual branches separately, then we use SDG as the guidance of visual feature representations for saliency and fuse it with PVT-SOD at multi-scale.

Overall, the contributions of this paper are as follows:
\begin{itemize}
    \item To the best of our knowledge, we are the first to underscore the subjective nature of “salience" in SOD and accordingly introduce semantic distillation from the generated image caption as an extra modality to semantically guide SOD visual features. This covers the gap between the subjective nature of SOD and its expensive labeling, by leveraging large quantity but inexpensive captioning data to compensate for demographic bias. Our SDG-SOD can also better uncover inter-objects and object-to-environment saliency where previous methods struggle.
    \item We employ an efficient hierarchical Transformer backbone to transfer SOD into a sequence-to-sequence learning problem, which achieves better performance than the mainstream CNN model design.
    \item We relabel the saliency map based on the panoptic segmentation labels of MS COCO~\cite{lin2014microsoft} for saliency prediction. This minimizes the labeling cost and produces finer SOD results.
    \item Extensive experiments demonstrate the effectiveness of our \method~model, outperforming the state-of-the-art (SOTA) methods on DUTS, ECSSD, DUT-OMRON, HKU-IS, PASCAL-S datasets.
\end{itemize}

\begin{figure*}
    \centering
    \includegraphics[width=0.96\linewidth]{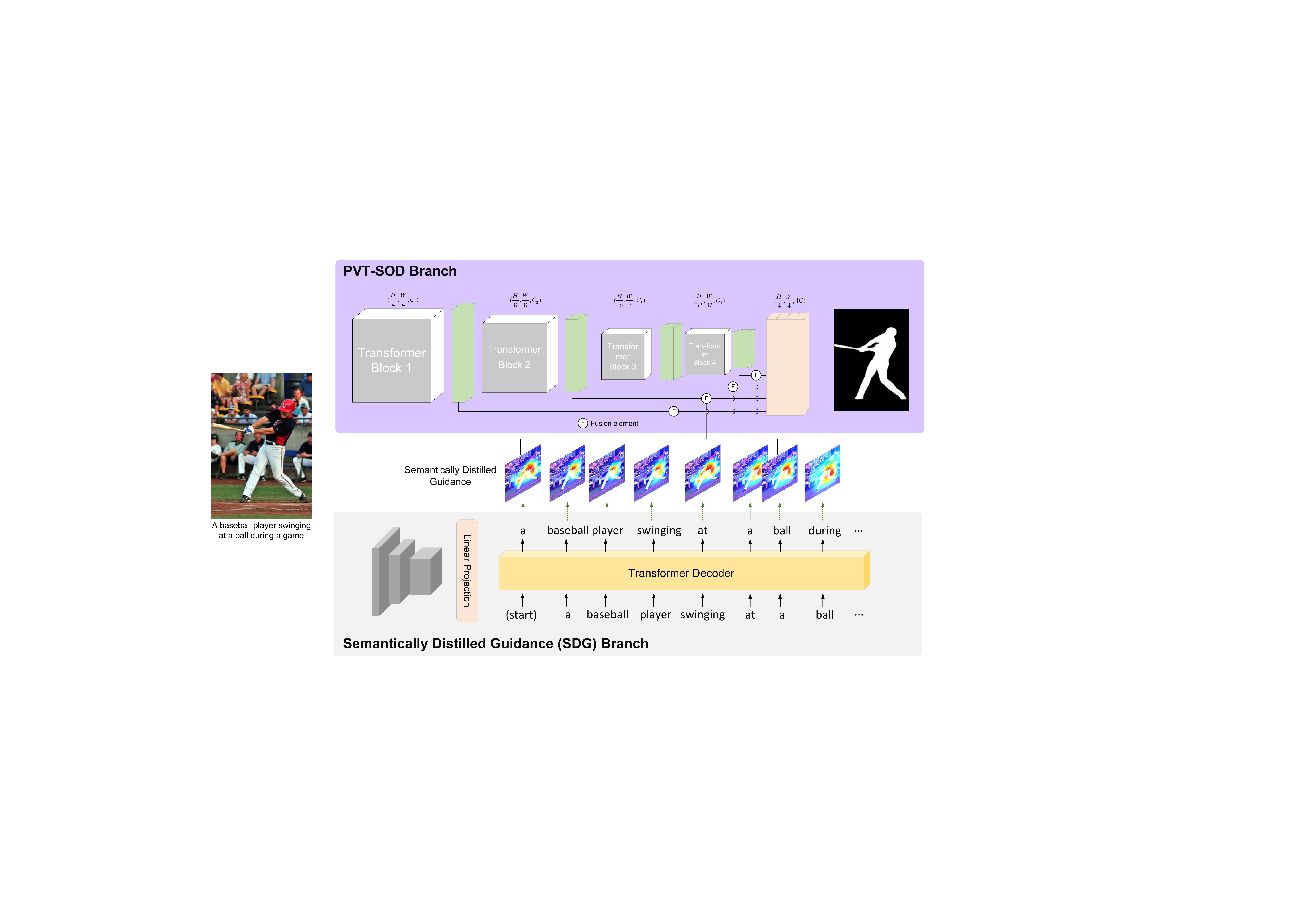}
    \caption{Overview of the proposed \method~network. The \method~network consists of two branches: Semantically distilled guidance (SDG) and PVT-based Salient Object Detection (PVT-SOD). We first train the dual branches separately, then we use SDG as the guidance of visual feature representations for saliency and fuse it with PVT-SOD at multi-scale.}
    \label{fig:pipeline}
\end{figure*}

\section{Related Works}\label{sec:related-works}
\subsection{Salient Object Detection}
    Early salient object detection methods are mainly based on hand-crafted features like color contrast, texture, and certain prior to extract saliency information, which focuses on low-level information. Recently, CNNs~\cite{qin2020u2} have been widely used to extract multi-level information to produce the saliency maps.
    
    Previous works focus on designing effective decoders to fuse multi-level features to obtain rich semantic information for accurate salient object detection.
    Wei \etal~\cite{wei2020f3net} proposed the F$^3$Net that consists of a cross-feature module and a cascaded feedback decoder to fuse multi-level image features. 
    Qin \etal~\cite{qin2019basnet} introduce a boundary-aware refinement network, which is capable of producing saliency maps with sharp boundaries.  
    Qin \etal~\cite{qin2020u2} further proposed a U$^2$-Net architecture to capture more contextual information from different scales.  
    Tang \etal~\cite{tang2021disentangled} disentangled the SOD task into a low-resolution saliency classification network and a high-resolution refinement network to capture sufficient semantics at low-resolution and refine the saliency value of pixels at high-resolution.
    Zhang \etal~\cite{zhang2021auto} designed a search cell to automatically decide multi-scale features aggregation for SOD and proposed a progressive polishing loss to further obtain exquisite boundaries.
    Ke \etal~\cite{ke2022recursive} proposed a contour-saliency blending module to exchange information between contour and saliency and adopted recursive CNN to increase contour-saliency fusion.

    Existing methods on salient object detection are mainly CNN-based models that are texture-biased and single object-oriented, thus leading to imperfect performance on multi-objects and noisy scenarios. Semantic information guidance is required to boost SOD performance.


\subsection{Image Captioning.}
    The problem of generating natural language descriptions from visual data has long been studied in computer vision. 
    Early methods use pre-defined templates, such as object detector and attribute predictor to generate captions~\cite{socher2010connecting,yao2010i2t}.
    With the rise of Deep Learning based networks, RNNs~\cite{rennie2017self,vinyals2016show} are adopted as language models to decode corresponding visual features. 
    
    Due to the wide success of transformers on natural language processing and multi-media, image caption methods use transformers to either generate captions directly or fuse visual and language features.
    Herdade~\etal~\cite{herdade2019image} propose a object relation transformer and build image captions based on inter-object relations.
    Liu~\etal~\cite{li2019entangled} introduce a enTangled attention-based transformer that simultaneously exploits visual and semantic information.
    Huang~\etal~\cite{huang2019attention} propose an attention model that first generates an information vector and an attention gate, and then add another attention using element-wise multiplication to aggregate attended features.
    
    Recent works have demonstrated that image captions can guide the feature learning of various visual tasks.
    Karan Desai~\etal~\cite{desai2021virtex} propose a pre-train approach using semantic dense captions to learn visual representations.
    We believe distilled semantic caption information can guide the feature learning of salient object detection tasks.
    
\subsection{Transformers on Dense Prediction.}
    Due to the recent success of transformer on vision tasks~\cite{zheng2021rethinking,wang2021pyramid}, there has been a surge of interest to introduce Transformers to dense prediction tasks like semantic segmentation.
    Zheng~\etal~\cite{zheng2021rethinking} propose SETR, which adopts ViT as backbone and test reconstruction results with different CNN decoders. 
    Wang~\etal~\cite{wang2021pyramid} proposed a pyramid vision transformer(PVT), showing promising results on semantic segmentation. 
    A Significant amount of efforts have been made to improve the performance of the transformer on dense prediction tasks, such as Swin~\cite{liu2021swin}, Twins~\cite{chu2021twins}, yet pixel-wise dense prediction still remains a difficult task for transformers.
\section{Methodology}
\label{sub:methodology}

In this section, we describe the SDG-SOD network, which consists of two branches: the semantically distilled guidance (SDG) and the PVT-based salient object detection (PVT-SOD). First, we train the two branches separately, PVT-SOD for salient object detection and SDG for image caption generation. Then we use SDG as the guide for the visual feature representations of saliency and fuse it with PVT-SOD. The overall architecture of the SDG-SOD network is shown in Figure~\ref{fig:pipeline}.     

\subsection{PVT-SOD Branch}
\label{sub:transformer_encoder}

\subsubsection{Pyramid Vision Transformer}
\label{sub:pyramid_vision_transformer}
The purpose of the PVT-SOD branch is to obtain multi-level feature maps from the input image. Both coarse features with high resolution and fine-grained features with low resolution are required for meaningful segmentation of salient objects. Given an input image of size $H\times W\times 3$, we perform overlapping patch merging to gradually reduce the resolution of the feature map from $\frac{H}{4}\times\frac{W}{4}$ to $\frac{H}{32}\times\frac{W}{32}$. Each feature map $F_i$ generated by patch merging is fed to the efficient transformer block for feature learning.
 
\subsubsection{Efficient Self-Attention.}
\label{sub:self_attention}
To reduce the high computational complexity of multi-head self-attention, we use the sequence reduction process introduced in \cite{wang2021pyramid} to reduce the spatial scale of $K\in\mathbb{R}^{(H_{i}W_{i})\times C_{i}}$ and $V\in\mathbb{R}^{(H_{i}W_{i})\times C_{i}}$ in each head before the attention operation at stage-$i$:

\begin{equation}
\begin{aligned}
    &\hat{X}=Norm(Reshape(X, R_{i})W^{X})\\
    &W^{X}=Linear(C_{i}\cdot R_{i}, C_{i})
\end{aligned}
\end{equation}

where $X\in\mathbb{R}^{(H_{i}W_{i})\times C_{i}}$ is the input patch sequence. $Reshape(X, R_{i})$ refers to reshape $X$ to the one with a shape of $\frac{H_{i}W{i}}{R_{i}}\times (C_{i}\cdot R_{i})$. And then an MLP network $W^{X}$ is used to learn a mapping from $C_{i}\cdot R_{i}$ dimensions to $C_{i}$ dimensions. $Norm(\cdot)$ refers to layer normalization~\cite{ba2016layer}. Then the self-attention operation is conducted with the new $K\in\mathbb{R}^{(\frac{H_{i}W{i}}{R_{i}})\times C_{i}}$ and $V\in\mathbb{R}^{(\frac{H_{i}W{i}}{R_{i}})\times C_{i}}$: 
\begin{equation}
    Attention(Q,K,V)=Softmax(\frac{QK^T}{\sqrt[]{d_{head}}})V
\end{equation}

Therefore, the complexity of the self-attention mechanism is reduced by $R_{i}$ times. In our implementation, we set the reduction ratio $R_{i}$ to [64, 16, 4, 1] from stage-1 to stage-4. The PVT-SOD module outputs a list of multi-resolution feature maps for further semantically distilled guidance. 

\subsection{Semantically Distilled Guidance Branch}

From an input image $I$, we aim at generating an image caption and a list of attention maps $M_j, j=1, \ldots, T$ representing the visually highlighted region of each word in the corresponding image, where the caption has $T$ words in total.
The word-level attention maps are used to aggregate visual information of the PVT-SOD model at different semantic levels. 
There are two main steps: (a) generating an image caption from the input image; (b) retrieving the word-level attention maps for each word within the sentence.

Our semantically distilled guidance branch is built on a transformer-based image captioning model~\cite{desai2021virtex} to generate guided attention maps. Such architecture is more suitable to fuse grid visual features provided by the PVT-SOD branch, where each attention value can be applied on its corresponding visual patch. 
Our semantically distilled guidance branch contains two parts: a visual encoder and a Transformer-based textual decoder. 
The input of the visual encoder is the same input image $I$ for the PVT-SOD. Then, the Transformer-based textual decoder decodes the visual features and generates a corresponding image caption $C = (c_0, c_1, ..., c_T, c_{T+1})$. The start of the image caption is $c_0 = [SOS]$,while $c_{T+1} = [EOS]$ indicates the end of a caption sequence.

\textbf{Visual Encoder:} The visual encoder uses a convolutional network to compute the downsampled visual features. For the input image $I$, we use the ResNet-50~\cite{he2016deep} as the visual encoder to extract grid feature $C\in\mathbb{R}^{2048\times(7\times 7)}$, followed by a linear projection layer before sending it to the textual decoder.

\textbf{Textual decoder:} The textual decoder receives a set of grid visual features and outputs the corresponding image caption and per-word visual attention maps for semantic guidance to help regularize the feature learning of salient object detection. We predict the captions in both forward and backward manner and utilize Transformer\cite{vaswani2017attention} as the backbone of textual decoder, which adopts a self-attention and cross attention mechanism to fuse visual features using textual queries.

The inputs of the textual decoder module are a set of image features from the visual encoder and a list of caption tokens.
Grid visual features fed into the textual decoder are tokenized to a sequence of patch features $G\in\mathbb{R}^{D_I\times N_I}$, where each $N_I = 7\times 7$ patch has a feature vector with $D_I-$dimension.   

The first tokens $c_0 = [SOS]$ indicates the start of the sentence. The transformer backbone iteratively predicts each word in the caption sentence. The prediction ends when transformer output $C_{T+1} = [EOS]$ label.

We sample the word-level attention map $M\in\mathbb{R}^{7\times7}$ in the cross attention module within the transformer decoder, which finds the highlighted visual regions corresponding to each word token, and use a bilinear upsample to resize the attention map to match the size of each feature map.
The above attention maps are further used as semantically distilled information to help regularize visual feature learning of the PVT-SOD branch.

\subsection{Semantic Distillation Guided SOD Decoder}

PVT-SOD backbone extracts multiple features $F_i$ at different semantic and resolution levels. 
We argue that features at different semantic levels can be aggregated using semantic distillation guided information. 
The word-level caption attention map obtained from the semantically distilled guidance branch contains semantically distilled information on image saliency, especially when multiple objects occur in the image. 
For instance, the attention maps corresponding to the subject and object of the caption sequence can aggregate the visual clues for different saliency objects, and the attention maps for predicate and preposition can highlight the semantic correlation among different saliency objects.

The word-level attention maps $M_j$ output by the image caption guidance branch and the visual feature maps $F_i, i\in[1, 4]$ from multi-stage of PVT-SOD are fused at multi-scale levels, then fed into the semantic distillation guided SOD decoder. Considering that different word focuses on specific region in the semantically distilled salient information, we believe that collecting all attention maps for SOD guidance can lead to better performance. 

We resize the attention maps $M_j$ to match the resolution of the feature map from PVT-SOD at each scale, and use the element-wise dot product to highlight the corresponding regions of the visual feature map as follows: 
\begin{equation}
    \bar{F_i} = \xi(F_i\cdot normalize(\sum_{j=1}^{T}Resize(M_j))) + F_i ,\forall~i
\end{equation}

where $T$ is the length of the caption generated from the semantically distilled guidance branch. We sum all the resized word-level attention maps and element-wise multiply them by $F_i$ after normalization. A linear projection layer $\xi$ is followed to facilitate feature fusion. The projected features are added to the original visual feature $F_i$ for a residual connection.

Then the aggregated multi-level feature maps $\bar{F_i}$ are aligned to the same feature dimension using MLP layers. The output feature maps are resized to 1/4 of the original image resolution and concatenated. An additional FC layer is followed to predict the final saliency map.
\section{Experiment}
\label{sec:experiment}

\subsection{Datasets}

\noindent\textbf{Pre-training Dataset:} Recent commonly-used SOD training datasets~\cite{wang2017learning} are data-insufficient due to the high cost of manual pixel-wise saliency map annotation. To address this issue, we propose our novel COCO annotations (\dataset) by relabeling the saliency map based on the panoptic segmentation labels of MS COCO~\cite{lin2014microsoft}. The~\dataset~contains more than 35000 training images, which is much larger than the existing SOD dataset. More details about~\dataset~are described in Appendix~\ref{coco_Annotation}. 

\noindent\textbf{Training Datasets:} We fine-tune our network on the \textbf{DUTS-TR} dataset. \textbf{DUTS-TR} is currently the largest dataset for salient object detection with accurate pixel-wise annotations and contains a total of 10553 images.

\noindent\textbf{Evaluation Datasets:} We evaluate our models on five frequently used benchmark datasets: DUT-OMRON\cite{yang2013saliency}, DUTS-TE\cite{wang2017learning}, HKU-IS\cite{li2015visual}, ECSSD\cite{yan2013hierarchical}, PASCAL-S\cite{li2014secrets}.

\begin{table*}[h]
    \centering
    \scalebox{0.9}{
    \begin{tabular}{c|c|c|c|c|c}
    \hline
     & DUTS-TE\cite{wang2017learning} & ECSSD\cite{yan2013hierarchical} & DUT-OMRON\cite{yang2013saliency} & HKU-IS\cite{li2015visual} & PASCAL-S\cite{li2014secrets} \\
    Method & $S_{\alpha}\uparrow$ $F_{\beta}\uparrow$ $E_{\xi}\uparrow$ $M\downarrow$ & $S_{\alpha}\uparrow$ $F_{\beta}\uparrow$ $E_{\xi}\uparrow$ $M\downarrow$ & $S_{\alpha}\uparrow$ $F_{\beta}\uparrow$ $E_{\xi}\uparrow$ $M\downarrow$ & $S_{\alpha}\uparrow$ $F_{\beta}\uparrow$ $E_{\xi}\uparrow$ $M\downarrow$ & $S_{\alpha}\uparrow$ $F_{\beta}\uparrow$ $E_{\xi}\uparrow$ $M\downarrow$ \\
    \hline
    CPD\cite{wu2019cascaded} & .869 .821 .898 .043 & .913 .909 .937 .040 & .825 .742 .847 .056 & .906 .892 .938 .034 & .848 .819 .882 .071 \\
    SCRN\cite{wu2019stacked} & .885 .833 .900 .040 & .920 .910 .933 .041 & .837 .749 .847 .056 & .916 .894 .935 .034 & .869 .833 .892 .063 \\
    PoolNet\cite{liu2019simple} & .887 .840 .910 .037 & .919 .913 .938 .038 & .831 .748 .848 .054 & .919 .903 .945 .030 & .865 .835 .896 .065 \\
    BASNet\cite{qin2019basnet} & .876 .823 .896 .048 & .910 .913 .938 .040 & .836 .767 .865 .057 & .909 .903 .943 .032 & .838 .818 .879 .076 \\
    U$^{2}$Net\cite{zhou2020interactive} & .873 .826 .896 .044 & .928 .923 .947 .033 & .847 .776 .867 .054 & .916 .903 .943 .031 & .844 .822 .873 .074 \\
    EGNet\cite{zhao2019egnet} & .878 .824 .898 .043 & .914 .906 .933 .043 & .840 .755 .855 .054 & .917 .900 .943 .031 & .852 .823 .881 .074 \\
    F$^{3}$Net\cite{wei2020f3net} & .888 .852 .920 .035 & .919 .921 .943 .036 & .839 .766 .864 .053 & .917 .910 .952 .028 & .861 .835 .898 .062 \\
    ITSD\cite{zhou2020interactive} & .886 .841 .917 .039 & .920 .916 .943 .037 & .842 .767 .867 .056 & .921 .906 .950 .030 & .860 .830 .894 .066\\
    RCSB\cite{ke2022recursive} & .881 .855 .903 .034 & .922 .927 .923 .033 & .835 .773 .855 .045 & \ .919 .923 .954 .027 & .860 .842 .852 .058 \\
    DCN\cite{wu2021decomposition} & .892 .859 .927 .035 & .928 .931 .955 .032 & .845 .779 .878 .051 & .922 .916 .958 .027 & .862 .845 .901 .062 \\
    HQSOD\cite{tang2021disentangled} & .892 .876 .931 .031 & .926 .941 .956 .030 & .843 .791 .876 .048 & .922 {\bf .929} .960 .026 & .859 .852 .905 .060 \\
    MSFNet\cite{zhang2021auto}& .877 .856 .927 .034 & .915 .929 .951 .033 & .832 .778 .873 .050 & .908 .914 .956 .027 & .852 .850 .901 .061\\
    \hline
    PVT-SOD & .898 .865 .933 .030 & .933 .933 .958 .028 & .855 .791 .883 .044 & .921 .918 .960 .026 & .866 .851 .906 .057 \\
    PVT-SOD$+$ & .903 .873 .938 .028 & .937 .940 .964 .025 & .865 .811 .902 .043 & .922 .918 .958 .026 & .869 .855 .911 .055  \\
    \method & .905 .878 .943 .027 & .935 .940 .964 .025 & .865 .806 .898 .043 & .927 .924 .962 .024 & .870 .858 .913 .054 \\
    \method $+$ & \textbf{.911} \textbf{.889} \textbf{.946} \textbf{.026} & \textbf{.938} \textbf{.943} \textbf{.964} \bf .024 & \textbf{.865} \textbf{.817} \textbf{.905} \textbf{.042} & \textbf{.927} .926 \textbf{.962} \textbf{.024} & \textbf{.870} \textbf{.861} \textbf{.913} \textbf{.053} \\
    \hline
    \end{tabular}}
    \caption{Performance comparison with benchmark RGB salient object detection models. PVT-SOD$+$: our PVT-SOD model with \dataset~pre-training. \method $+$: our \method~model with \dataset~pre-training on the PVT-SOD branch.}
    \label{tab:performance_comparison}
\vspace{-6pt}
\end{table*}

\begin{figure}[t]
    \centering
    \includegraphics[width=0.98\linewidth]{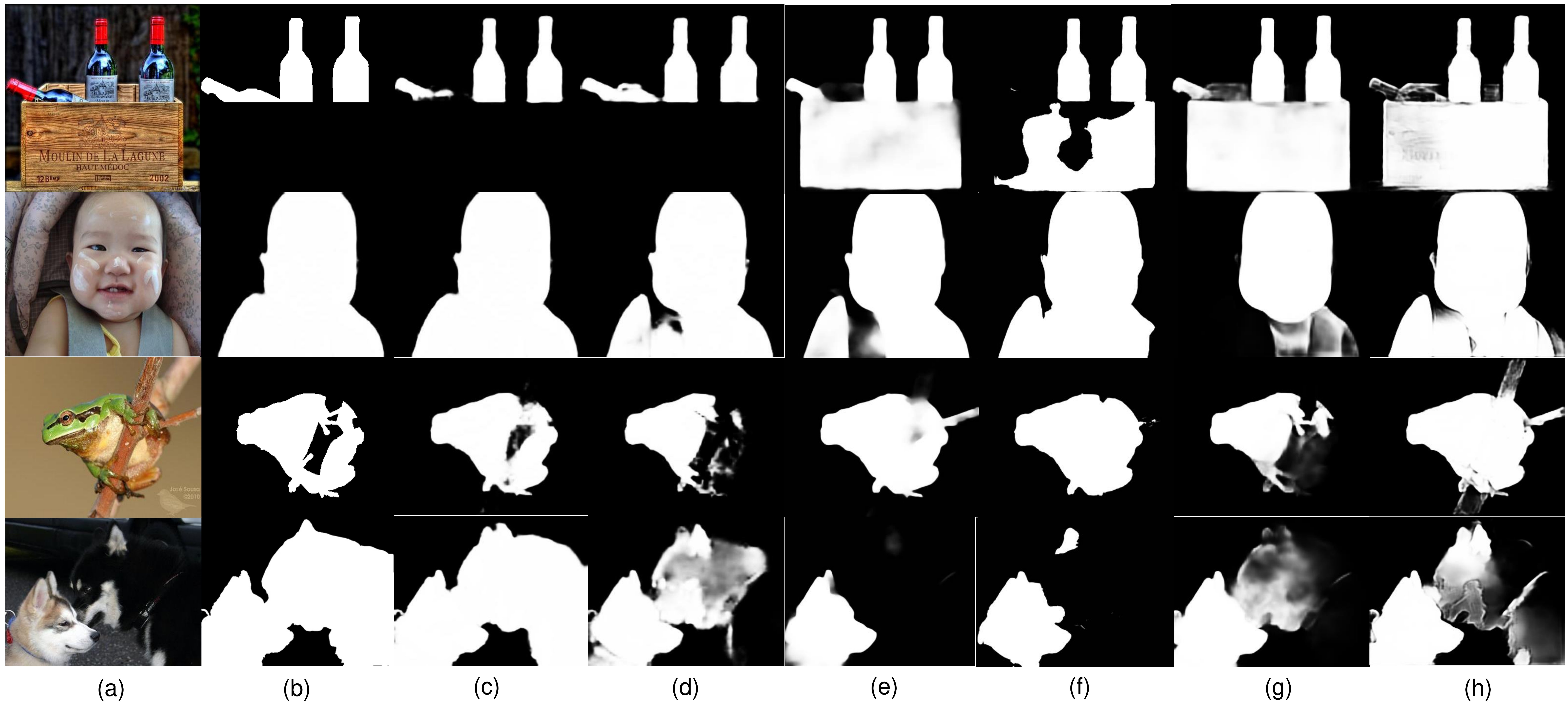}
    \caption{Visual comparison with the state-of-the-art RGB salient object detection models and our PVT-SOD~network model: (a) image, (b) GT, (c) PVT-SOD with \dataset~pre-training, (d) PVT-SOD w/o \dataset~pre-training, (e) DCN\cite{wu2021decomposition}, (f) HQSOD\cite{tang2021disentangled}, (g) F$^{3}$Net\cite{wei2020f3net}, (h) ITSD\cite{zhou2020interactive}.}
    \label{fig:visualization_pvt}
\vspace{-8pt}
\end{figure}

\subsection{Evaluation Metrics}
To evaluate the performance of our \method, we use four metrics to measure the performance. We report the Mean Absolute Error $M$, Mean F-measure $F_{\beta}$, Mean E-measure $E_{\xi}$ and S-measure $S_{\alpha}$. In addition, precision-recall curves and F-measure curves are drawn to show the whole performance. 

\textbf{PR curve} shows the precision-recall curves under five different datasets, which can evaluate the holistic performance of \method~and the state-of-the-art (SOTA) models.

\textbf{MAE} $M$ is denoted as the pixel-wise difference between predicted $P$ and ground-truth saliency map $G$.
\begin{equation}
    MAE = \frac{1}{H\times W}\sum_{i=1}^{H}\sum_{j=1}^{W}|P(i, j) -G(i, j)|
\end{equation}
where $H$ and $W$ are the height and width of the GT saliency maps. 

\textbf{F-measure} $F_{\beta}$ is computed to have a comprehensive measure on both precision and recall.
\begin{equation}
    F_{\beta} = \frac{(1 + \beta^2)\times Precision \times Recall}{\beta^2\times Precision + Recall}
\end{equation}
where $\beta^2$ is set to 0.3 and we report the mean $F_{\beta}$ and the F-measure curves following previous works\cite{wei2020f3net}.

\textbf{E-measure} $E_{\xi}$ is proposed to capture both image-level statistics and local pixel matching information.

\textbf{S-measure} is used to measure the structural similarity of the predicted and GT saliency maps, which is a weighted sum of region-aware ($S_r$) and object-aware ($S_o$) structural similarity.

\begin{equation}
    S_{\alpha} = (1 - \alpha)S_{r} + \alpha S_{0}
\end{equation}

\subsection{Implementation Details}
The transformer backbone\cite{xie2021segformer} of PVT-SOD is initialized with weights trained on semantic segmentation task. We resize all training images to size $352\times 352$. We pre-train our \method~on \dataset~for 30 epochs and fine-tune it on the DUTS dataset for 80 epochs with an initial learning rate of $5\times10^{-5}$. We follow previous work and adopt the loss term as described in F$^3$Net\cite{wei2020f3net}.

\begin{figure*}[t]
    \centering
    \includegraphics[width=1.0\linewidth]{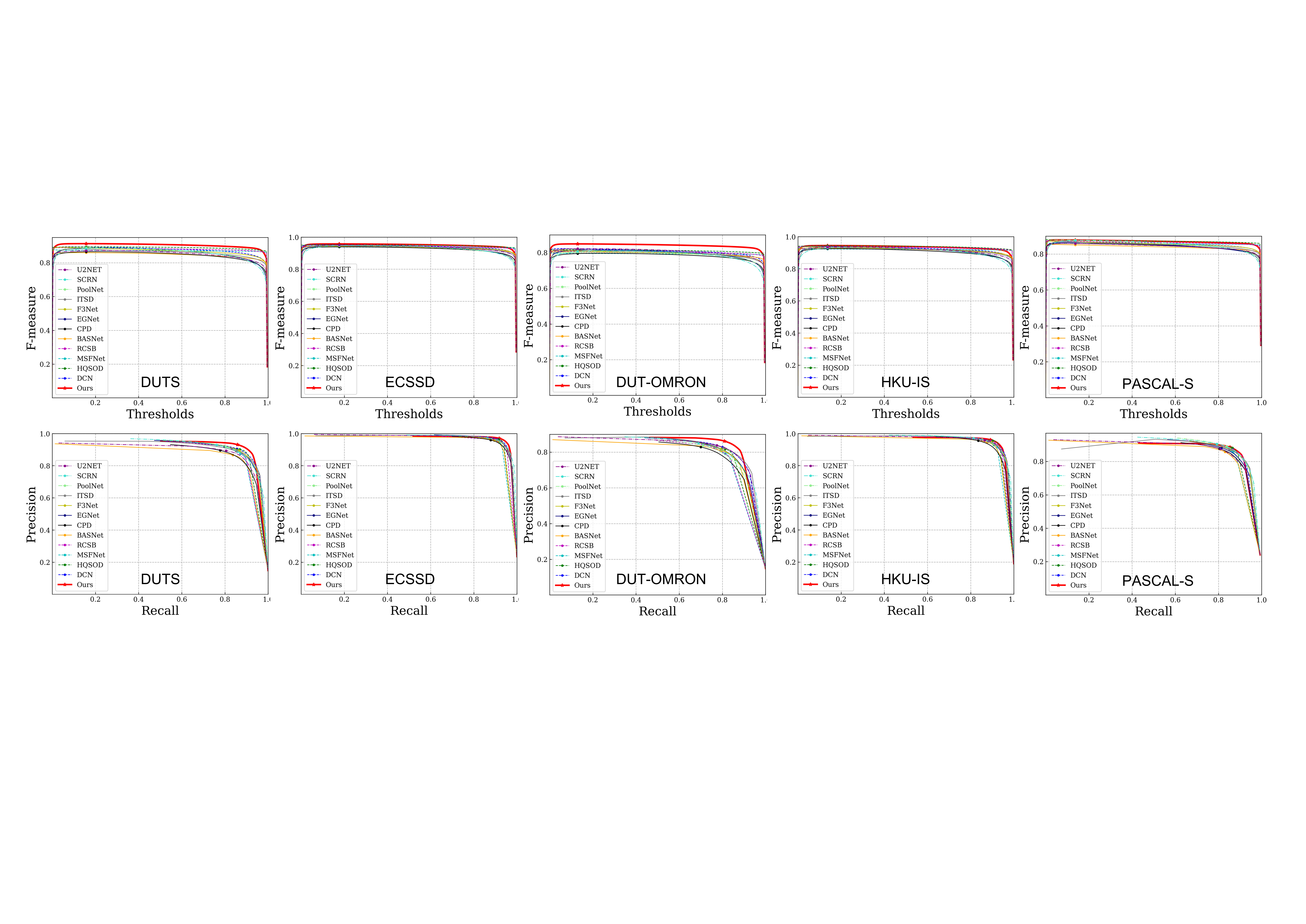}
    \caption{Illustration of F-measure curves (the first row) and PR curves (the second row) on the five largest datasets.}
    \label{fig:pr_curve}
    \vspace{-15pt}
\end{figure*}

\subsection{Comparative Study with the State-of-the-arts}

To prove the effectiveness of our proposed \method~method and each of its elements, we compare our models with 12 current state-of-the-art models: CPD\cite{wu2019cascaded}, SCRN\cite{wu2019stacked}, PoolNet\cite{liu2019simple}, BASNet\cite{qin2019basnet}, U$^2$Net\cite{zhou2020interactive}, EGNet\cite{zhao2019egnet}, F$^{3}$Net\cite{wei2020f3net}, ITSD\cite{zhou2020interactive}, RCSB\cite{ke2022recursive}, DCN\cite{wu2021decomposition}, HQSOD\cite{tang2021disentangled} and MSFNet\cite{zhang2021auto}. For fair comparison, we either use saliency maps provided by the authors or run their released codes.

\textbf{PVT-SOD.} As shown in Table~\ref{tab:performance_comparison}, our PVT-SOD shows significant superiority over almost all the competing methods across five datasets with respect to four metrics, and is better than HQSOD on $S_{\alpha}$, $E_{\xi}$ and $M$ metrics, while slightly worse than it on $F_{\beta}$ metric of 4/5 datasets. Especially, in the four metrics of the DUT-OMRON dataset, our method is completely ahead of HQSOD. The quantitative comparison demonstrates that the PVT backbone is more advantageous than CNN in extracting visual feature representations for pixel-level dense prediction of SOD. The comparative visualizations in Figure~\ref{fig:visualization_pvt} also show that our PVT-based network can effectively integrate global understanding and reduce the structural loss for better salient detection, even when the object is occluded. It leads to a balance between local texture details and salient structural integrity, benefiting from the global receptive field provided by the global attention mechanism of PVT. 

\textbf{Pre-training on \dataset.} We also report the PVT-SOD and \method~results with and w/o \dataset~pre-training. PVT-SOD$+$ and \method$+$~denote separately the results of PVT-SOD and \method~ pre-trained by \dataset. The results demonstrate that pre-training on \dataset~improves the performance of PVT-SOD both with and w/o semantically distilled guidance on all datasets except HKU-IS datasets. This indicates that our \dataset~can largely eliminate the overfitting problem caused by insufficient training data.

\textbf{Semantically distilled guidance.} After applying semantically distilled guidance, our~\method~further improves the effectiveness of salient object detection. As shown in Table~\ref{tab:performance_comparison}, our proposed model highlights its leading position on all four metrics across 4/5 datasets, except for minor regression compared to HQSOD~\cite{tang2021disentangled} in $ F_{\beta}$ of the HKU-IS dataset. Considering all the scores, our work achieves the state-of-the-art performances in the task of the RGB salient object detection. We also show the precision-recall curves (PR) and the F-measure curves in Figure~\ref{fig:pr_curve} to evaluate the holistic performance of models. The curves are consistent with the results in Table~\ref{tab:performance_comparison} and demonstrate the significant advantages of our SDG-SOD in terms of both salient regions detection and pixel prediction accuracy. Benefiting from the advantage of image captioning in salient semantic guidance and effective fusion with PVT-SOD at multi-scale, our \method~achieves a better performance in saliency detection. Moreover, our method also provides an alternative perspective on pixel-level dense prediction tasks, where visual-to-textual model is used as a guide to regularize visual feature representations without heavily relying on expensive pixel annotations.  

\begin{figure*}[t]
    \centering
    \includegraphics[width=1.0\linewidth]{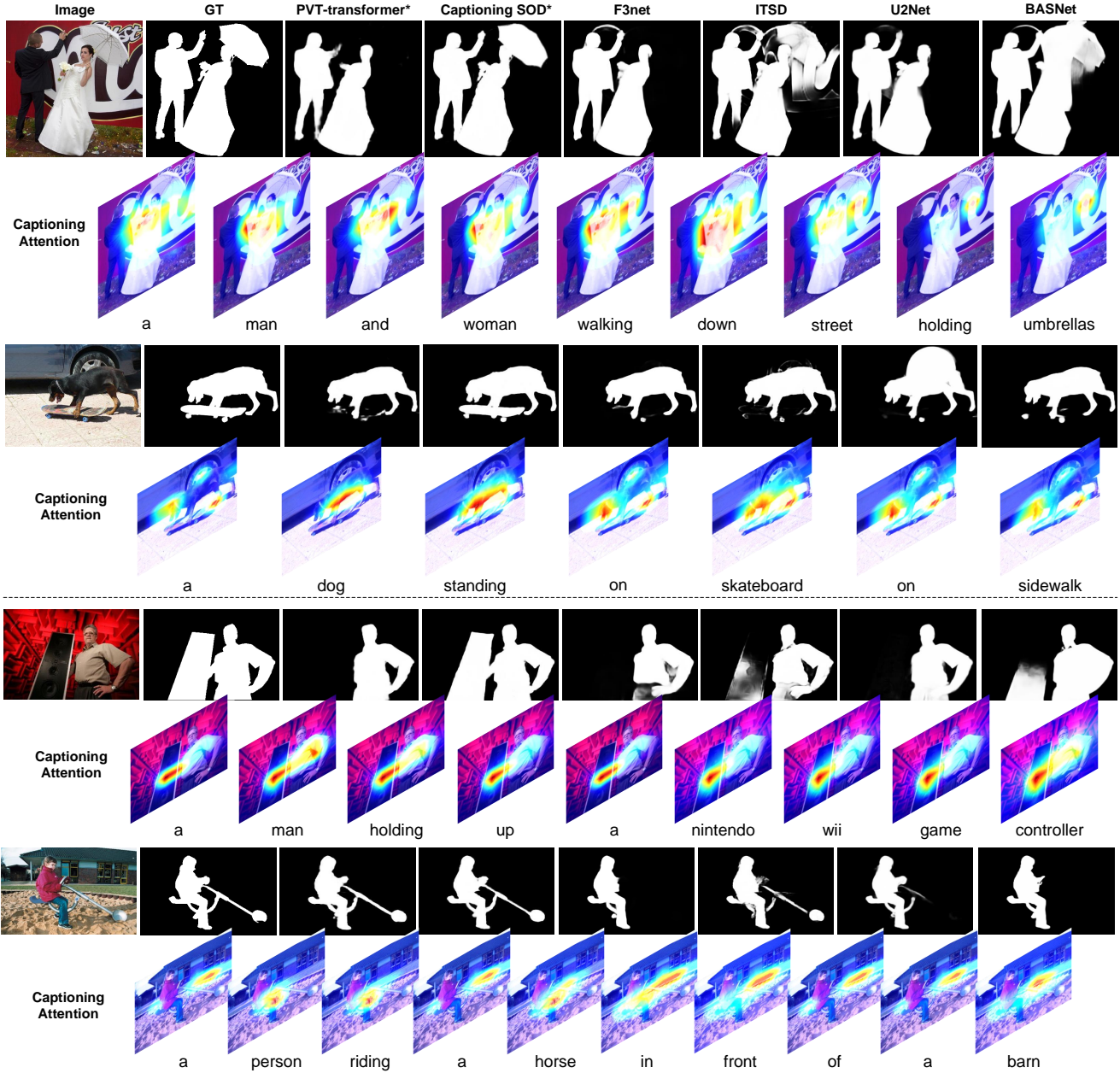}
    \caption{Comparison results with the state-of-the-art RGB salient object detection models and our PVT-SOD model and \method~model. $^{*}$ means our proposed models.}
    \label{fig:visualization}
\end{figure*}

\textbf{Visual comparison.} 
To give an intuitive understanding of the significance of the semantically distilled guidance, we visualize sample results of our \method~and other SOTA methods along with the sequence of word-level attention maps on DUTS-TE datasets in Figure~\ref{fig:visualization}. Compared to traditional CNN-based methods~\cite{wei2020f3net,qin2019basnet,qin2020u2,zhou2020interactive,wu2021decomposition, tang2021disentangled} which are demographically biased and single-object-oriented, our \method~can utilize a caption-guided attention map to highlight different semantic object regions. Examples are in Figure~\ref{fig:visualization}, where subjects and objects highlight different salient objects (\eg ‘man', ‘woman' and ‘umbrella' in Row 1, ‘dog' and ‘skateboard' in Row 2), and predicates highlight the semantic correlation among salient objects (\eg ‘holding' in Row 1 and ‘standing on' in Row 2). These characteristics can help our \method~ to achieve higher level of completeness in terms of saliency in an image, where other methods fail. More visual comparison details are described in Appendix~\ref{more_visual}.

Challengingly, the correlation between objects varies from scene to scene. One example is in Figure~\ref{fig:introduction} (b), distant from the person, the saliency of the unattended luggage is uncertain. The semantic association between them may not be accurately captured by relying solely on the larger visual perceptive field. However, with the help of semantically distilled guidance, the luggage is effectively detected to be salient based on the scene (\ie the airport) and contextual information. Visualization of the attention maps in Figure~\ref{fig:introduction} (b) also illustrates this. 

We note that despite our captioning model may sometimes generate imperfect captions (Row 3 and 4 in Figure~\ref{fig:visualization}) due to limited vocabulary or rare object classes, the generated
captioning attention map can still accurately highlight the salient visual feature regions and object correlations, which proves its attentive robustness. 

\begin{table}[t]
    \centering
    \scalebox{0.76}{
    \begin{tabular}{ccccccccc}
        \hline
        \multirow{2}{*}{Layers} & \multicolumn{4}{c}{w/o SDG} & \multicolumn{4}{c}{with SDG} \\
        \cmidrule(l){2-5} \cmidrule(l){6-9}
        &$S_{\alpha}\uparrow$ & $F_{\beta}\uparrow$ & $E_{\xi}\uparrow$ & $M\downarrow$ & $S_{\alpha}\uparrow$ & $F_{\beta}\uparrow$ & $E_{\xi}\uparrow$ & $M\downarrow$ \\
        \hline
        $[2, 2, 2, 2]$ & .871 & .825 & .906 & .045 & .869 & .849 & .911 & .037 \\
        $[3, 4, 6, 3]$ & .889 & .851 & .923 & .036 & .890 & .871 & .925 & .031 \\
        $[3, 4, 18, 3]$ & .898 & .865 & .933 & .030 & .905 & .878 & .943 & .027 \\
        \hline
    \end{tabular}}
    \caption{Ablation study of Transformer encoder on DUTS-TE\cite{wang2017learning} with and w/o the caption guidance.}
    \label{tab:transformer_backbone}
\vspace{-6pt}
\end{table}

\begin{table}[t]
    \centering
    \scalebox{0.90}{
    \begin{tabular}{ccccc}
        \hline
        Resolution&$S_{\alpha}\uparrow$ & $F_{\beta}\uparrow$ & $E_{\xi}\uparrow$ & $M\downarrow$\\
        \hline
        $[H/4, W/4]$ & .910 & .888 & .946 & .027 \\
        $[H/8, W/8]$ & .909 & .888 & .945 & .027 \\
        $[H/16, W/16]$ & .908 & .886 & .943 & .028 \\
        $[H/32, W/32]$ & .907 & .885 & .943 & .028\\
        \hline
    \end{tabular}}
    \caption{Impact of semantic distilled guidance on different SOD feature resolutions.}
    \label{tab:fusion_solution}
\end{table}

\begin{table}[t]
    \centering
    \scalebox{0.9}{
    \begin{tabular}{c|c|c|c|c}
        \hline
        Decoder & $S_{\alpha}\uparrow$ & $F_{\beta}\uparrow$ & $E_{\xi}\uparrow$ & $M\downarrow$ \\
        \hline
        PSP\cite{zhao2017pyramid} & .908 & .883 & .942 & .027 \\
        ASPP\cite{chen2017deeplab} & .910 & .885 & .944 & .026 \\
        APCNet\cite{he2019adaptive} & .910 & .881 & .942 & .028 \\
        ANNNet\cite{zhu2019asymmetric} & .876 & .821 & .923 & .040 \\
        \hline
        Ours & \textbf{.911} & \textbf{.889} & \textbf{.946} & \textbf{.026} \\
        \hline
    \end{tabular}}
    \caption{Ablation study of Transformer decoder on DUTE\cite{wang2017learning}.}
    \label{tab:decoder}
\vspace{-12pt}
\end{table}

\subsection{Ablation Study}
We validate the effectiveness of some key components used in our model. The ablation studies are conducted on the DUTS-TE dataset~\cite{wang2017learning}.

{\bf Semantically distilled guidance vs. pre-training:} We report the quantitative comparison results of our models with and w/o the semantically distilled guidance branch. As illustrated in Table~\ref{tab:performance_comparison}, the semantically distilled guidance improves the performance of our model on SOD, even better than our pre-training method, especially on the DUTS-TE, HKU-IS, PASCAL-S datasets. More advantageous, image captioning needs extremely low labeling costs.

{\bf Semantically distilled guidance on different PVT backbones:}
We test our \method~on four PVT block layer settings with and w/o semantically distilled guidance (SDG) to gain an overall understanding of the semantically distilled guidance effects on different transformer backbone sizes in PVT. The results in Table~\ref{tab:transformer_backbone} show that our semantically distilled guidance strategy significantly increase the performance of the model regardless of the backbone layer settings. We also notice that semantically distilled guidance can provide more semantic information to PVT-SOD for better performance on SOD, and the promotion effect on backbones with small sizes is more obvious.

{\bf Semantically distilled guidance on different SOD features:} 
We applied the captioning attention maps on four SOD feature scales from the PVT-SOD+ individually and evaluated each performance on the DUTS-TE dataset. The results in Table~\ref{tab:fusion_solution} show that SOD features with higher resolutions can be better guided by captioning knowledge. Together with Table 1, it reassures that multi-scale PVT fusion leads to stronger performance.

{\bf MLP decoder:} We then test the \method$+$ to evaluate the effectiveness of our MLP decoder, by comparing the results with two commonly used decoders PSP~\cite{zhao2017pyramid}, ASPP~\cite{chen2017deeplab}, APCNet~\cite{he2019adaptive} and ANNNet~\cite{zhu2019asymmetric} in Table~\ref{tab:decoder}. The results on $S_{\alpha}$, $F_{\beta}$, $E_{\xi}$, $M$ demonstrate the simplicity and effectiveness of an all-MLP based decoder architecture. 


\vspace{-5pt}
\section{Conclusion}
\vspace{-5pt}
\label{sec:conclusion}
We propose a semantically distillation guided SOD detection method \method~that produces accurate saliency maps by fusing semantic distilled knowledge from generated image captioning into the Visual-Transformer-based SOD framework. Our \method~can leverage explicit semantic information from caption to better cover demographic distribution without expensive labeling. Extensive experiments demonstrate the effectiveness of \method~and it outperforms SOTA algorithms on multiple datasets. For future work, we are going to propose an image captioning guided SOD data annotation method, aims at eliminating the individual variation on labeling. 



\appendix
  \renewcommand{\appendixname}{Appendix~\Alph{section}}

\begin{figure}[t]
    \centering
    \includegraphics[width=0.96\linewidth]{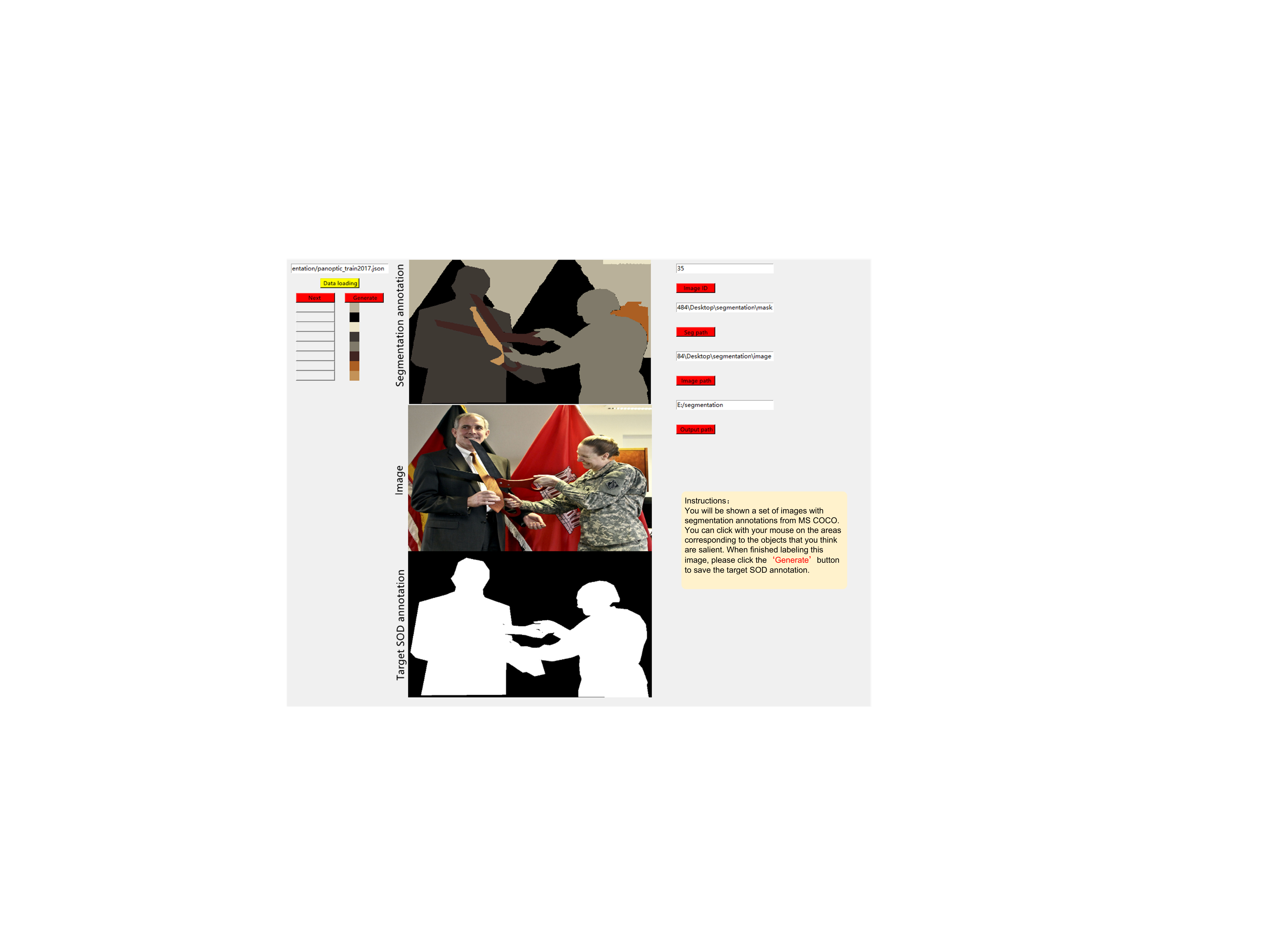}
    \caption{The UI for our relabeling tool. Annotators are shown the original image and segmentation annotation from MS COCO datasets. Then they can select salient objects by clicking on the corresponding areas in the segmentation annotation.}
    \label{fig:label_ui}
\end{figure}

\begin{figure}[t]
    \centering
    \includegraphics[width=0.96\linewidth]{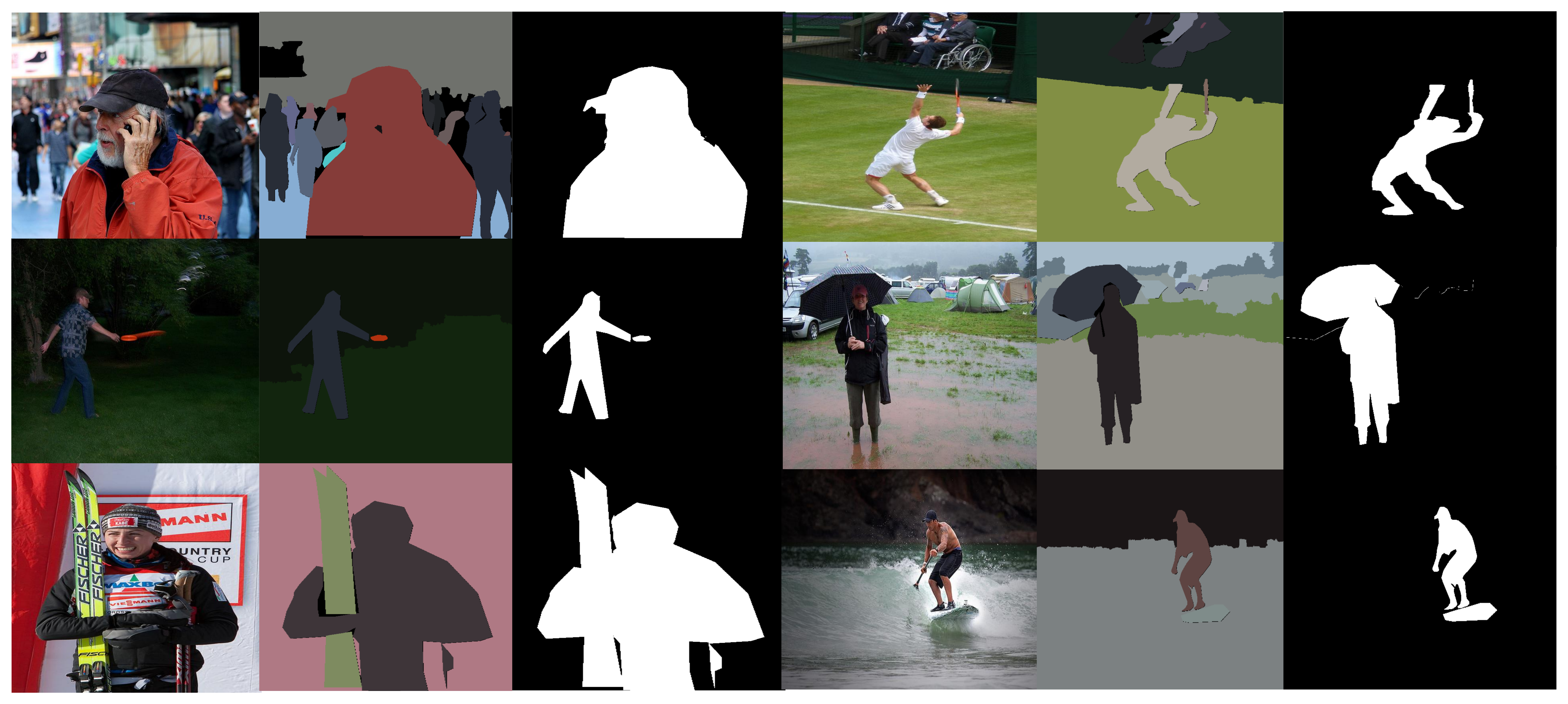}
    \caption{From left to right are our RGB images, panoptic labels and generated saliency maps.}
    \label{fig:coco}
\end{figure}

\section{COCO SOD Annotation.}\label{coco_Annotation}

%
Since the high-quality and pixel-wise annotations for the SOD task require large labeler populations, it is unrealistically expensive to acquire a
representative dataset that covers comprehensive and diverse scenes. Recent existing training sets for SOD are data-insufficient, \eg~the widely-adopted SOD training set \textbf{DUTS-TR}~\cite{wang2017learning}, contains a total of only 10553 images.

%
%


%
To address the above issue, we propose a pre-training strategy that we relabel the saliency annotations based on the panoptic segmentation labels of the MS COCO dataset. Figure~\ref{fig:label_ui} shows the tool UI for our relabeling, where annotators can manually select the panoptic segmentation labels for foreground objects that are considered as salient within the image. In Figure~\ref{fig:coco}, we show some examples in our \dataset. And then we labeled more than 35000 images with saliency map annotations. Unlike \textbf{DUTS-TR}, \dataset~provide saliency maps that have coarse annotations around object edges. 
Yet, we argue that pre-training on \dataset~can well erase the aforementioned overfitting problem.

\section{More Visualization Results.}\label{more_visual}

We visualize the qualitative results of our \method~and the state-of-the-art methods, as well as the generated captions and the corresponding word-level attention maps from our SDG branch on multi-object scenarios.
As shown in Figure~\ref{fig:visualization}, our SDG-SOD can utilize captioning-guided attention maps to highlight different salient objects and the correlation among them. This characteristic can help our SDG-SOD capture a higher level of completeness in terms of saliency under a limited amount of supervision and labeling effort, where other methods fail.

\begin{figure*}[t]
    \centering
    \includegraphics[width=1.05\linewidth]{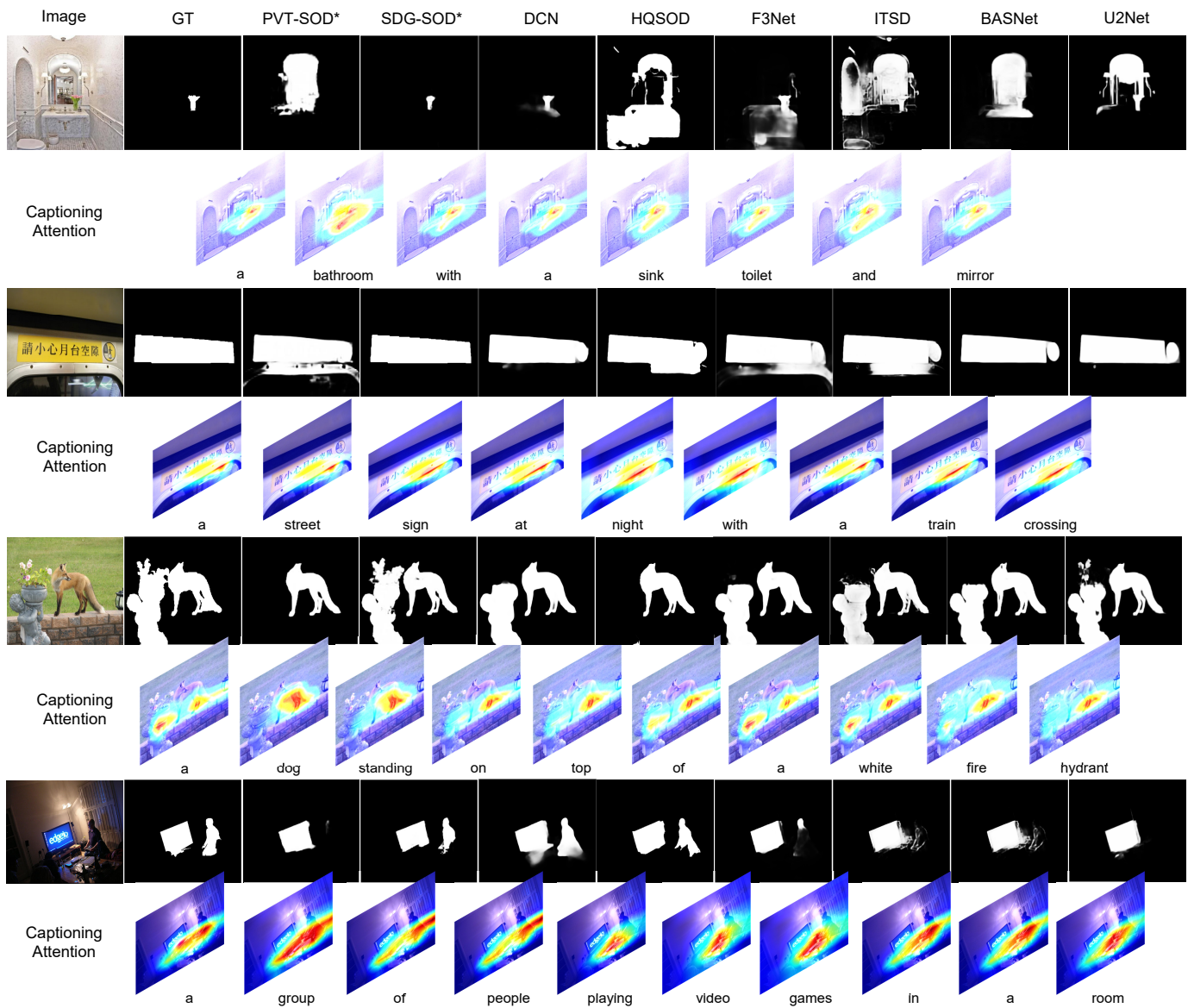}
    \caption{Visual comparisons of the state-of-the-art models and our proposed models on the DUT (Row 1), ECSSD (Row 2), HKU-IS (Row 3), PASCAL-S (Row 4) datasets.}
    \label{fig:more_visualization}
\end{figure*}

More visual comparisons on multiple datasets are shown in Figure~\ref{fig:visualization} and~\ref{fig:more_visualization}, which further demonstrate the effectiveness of our method. We argue that our \method~can severely improve the SOD performance, especially on multi-objects scenarios. 

%
%

\begin{figure}[t]
    \centering
    \includegraphics[width=\linewidth]{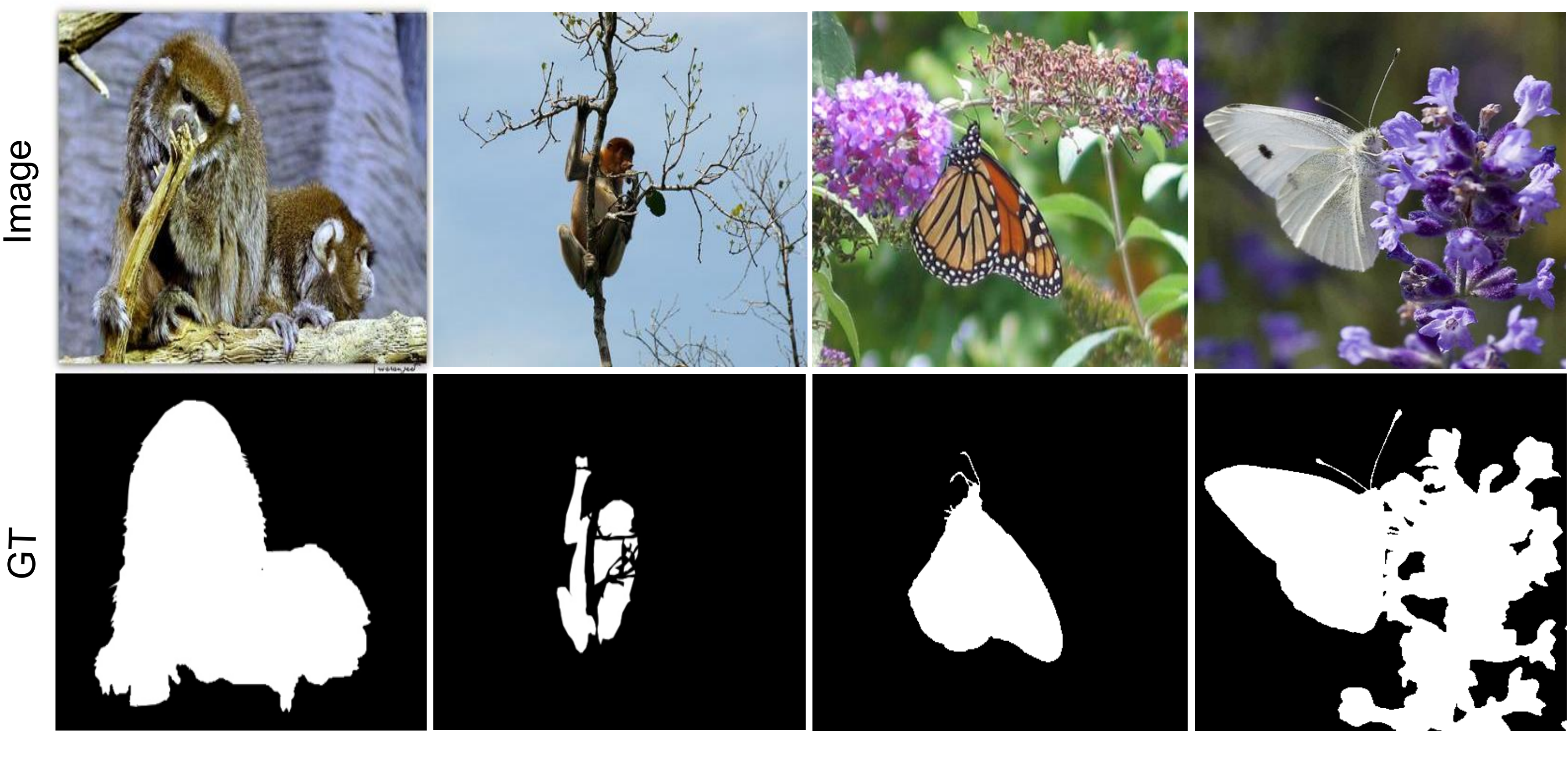}
    \caption{Visualization of some bias annotation examples within the \textbf{DUTS-TE}~\cite{wang2017learning} dataset.}
    \label{fig:my_label}
\vspace{-10pt}
\end{figure}

\section{Limitations.}\label{limitation}
Unlike some well-defined tasks such as object detection or segmentation, the definition of saliency in an image is highly subjective. If images with similar scenes are presented to a group of labelers, it's very natural that each individual may interpret the images a bit differently. As shown in Figure~\ref{fig:my_label}, the tree branch in front of a monkey (top left image) is considered to be salient, yet the branch (the top right image) in a similar position is considered as occlusion for the salient monkey.
The bottom row in Figure~\ref{fig:my_label} also shows the annotations of two similar images that both contain a butterfly and a flower. However, the bottom left saliency map contains only the butterfly, yet the bottom right considers the flower to be equally salient. Such variation may make it difficult for the SOD models to effectively learn the true distribution of saliency. 

To address this challenge, for future work we are going to propose an image captioning guided SOD data annotation method, aims at eliminating the individual variation on labeling.

%
%
%

\begin{figure*}[t]
    \centering
    \includegraphics[width=1\linewidth]{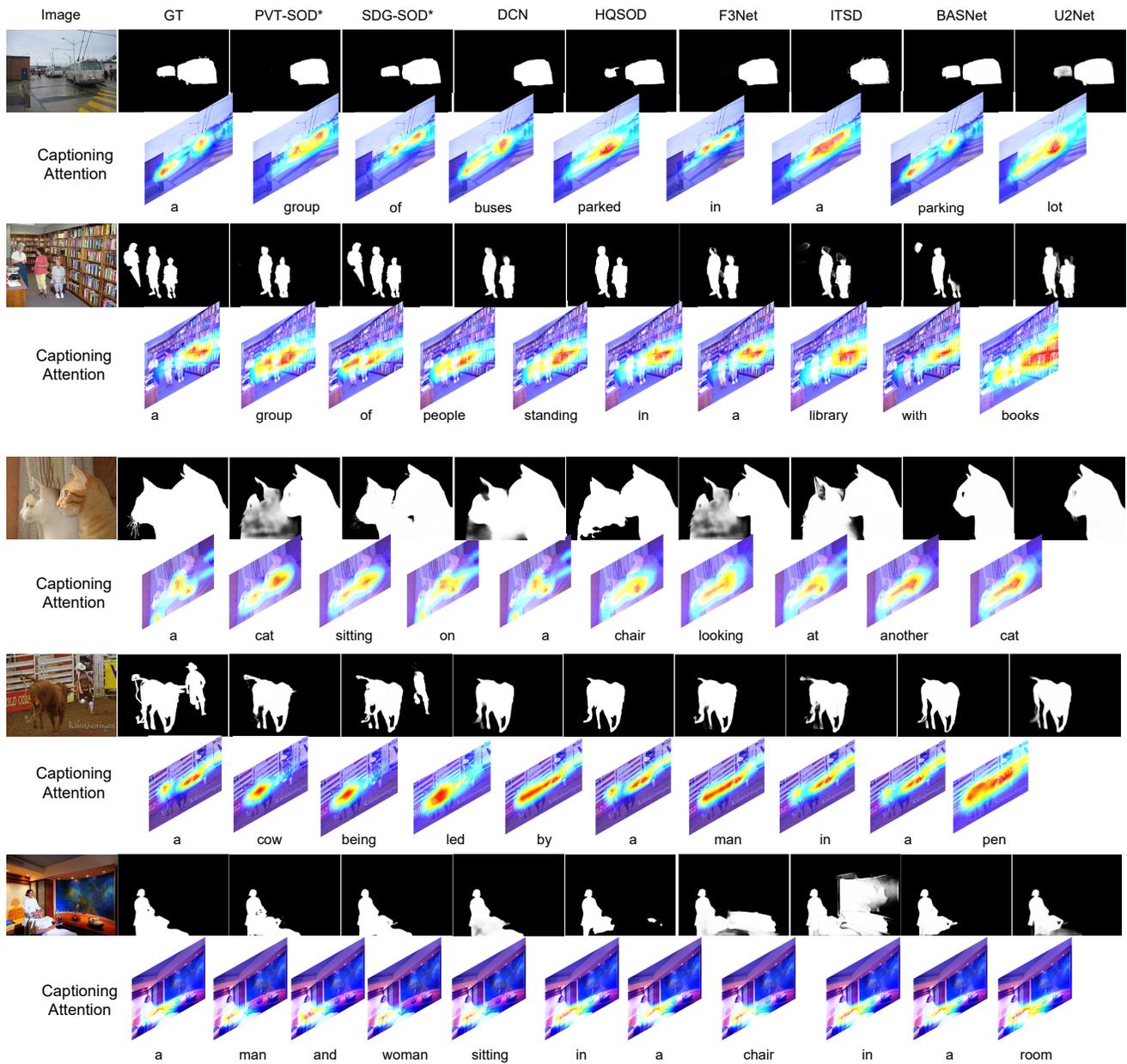}
    \caption{Visual comparisons of the state-of-the-art salient object detection models and our PVT-SOD and \method~models on the \textbf{DUTS-TE} dataset. * means our proposed models.}
    \label{fig:visualization}
\vspace{-14pt}
\end{figure*}

\end{document}